\documentclass[sigconf]{acmart}
\renewcommand\footnotetextcopyrightpermission[1]{} 
\usepackage{array}
\usepackage{caption}
\usepackage{subcaption}
\usepackage{multirow} 
\usepackage{algorithm}
\usepackage{placeins} 
\usepackage[noend]{algpseudocode}
\AtBeginDocument{%
  \providecommand\BibTeX{{%
    \normalfont B\kern-0.5em{\scshape i\kern-0.25em b}\kern-0.8em\TeX}}}

\setcopyright{none}
\acmDOI{}

\begin{document}
\pagestyle{plain}

\title{Improving the efficiency of GP-GOMEA for higher-arity operators}


\author{Thalea Schlender}
\email{t.schlender@lumc.nl}
\affiliation{%
  \institution{Leiden University Medical Center}
  \streetaddress{Albinusdreef 2}
  \city{Leiden}
  \country{The Netherlands}}

\author{Mafalda Malafaia}
\email{mafalda.malafaia@cwi.nl}
\affiliation{%
  \institution{Centrum Wiskunde \& Informatica}
  \streetaddress{..}
  \city{Amsterdam}
  \country{The Netherlands}}

\author{Tanja Alderliesten}
\email{t.alderliesten@lumc.nl}
\affiliation{%
  \institution{Leiden University Medical Center}
  \streetaddress{Albinusdreef 2}
  \city{Leiden}
  \country{The Netherlands}}

\author{Peter A.N. Bosman}
\email{peter.bosman@cwi.nl}
\affiliation{%
  \institution{Centrum Wiskunde \& Informatica}
  \city{Amsterdam}
  \country{The Netherlands}}
%\affiliation{
%\institution{Technical University Delft}
%\streetaddress{..}
%\city{Delft}
%\country{The Netherlands}
%}

\begin{abstract}
Deploying machine learning models into sensitive domains in our society requires these models to be explainable. Genetic Programming (GP) can offer a way to evolve inherently interpretable expressions. GP-GOMEA is a form of GP that has been found particularly effective at evolving expressions that are accurate yet of limited size and, thus, promote interpretability. Despite this strength, a limitation of GP-GOMEA is template-based. This negatively affects its scalability regarding the arity of operators that can be used, since with increasing operator arity, an increasingly large part of the template tends to go unused. In this paper, we therefore propose two enhancements to GP-GOMEA: (i) semantic subtree inheritance, which performs additional variation steps that consider the semantic context of a subtree, and (ii) greedy child selection, which explicitly considers parts of the template that in standard GP-GOMEA remain unused. 
We compare different versions of GP-GOMEA regarding search enhancements on a set of continuous and discontinuous regression problems, with varying tree depths and operator sets. Experimental results show that both proposed search enhancements have a generally positive impact on the performance of GP-GOMEA, especially when the set of operators to choose from is large and contains higher-arity operators.
\end{abstract}


\maketitle
\thispagestyle{empty}
\nocite{harrison2022gene, sijben2022multi}
\section{Introduction}
Over the last decades, machine learning models have continuously improved, many of which now achieve (beyond) human-level performance on a broad array of tasks. As a result, these models are embedded in an increasing number of areas in society. However, especially in sensitive domains such as medicine or judiciary, their application and deployment require trust. To ensure fairness and equality, models need to be interpretable and accountable, not only by preference but increasingly by law~\cite{vilone2020explainable, lipton2018mythos, harrison2022gene}. 

Interpretability is, however, challenging for many black-box machine-learning models used today. Due to this, explainable artificial intelligence (XAI) is increasingly gaining attention. 
In this field, inherently interpretable machine learning models are of special interest, as they do not attempt to explain a black box, but instead are white box models by design. 

One such inherently interpretable model is a mathematical expression (of comprehensible size and complexity) that describes the relationship between a given set of data points and their respective target values. The task of finding the best-fitting expression is called symbolic regression. Often, a predefined set of operators, variables, and constants is defined that may then be used in the expression. A popular search strategy is to evolve the expression via genetic programming (GP). 

Many classic GP approaches are prone to find accurate but large expressions, threatening their interpretability~\cite{virgolin2021improving}. To combat the continuous growth of expressions, which is also known as bloating, a fixed solution size can be used. GP-GOMEA~\cite{virgolin2017scalable, virgolin2021improving}, derived from the model-based Gene-pool Optimal Mixing Evolutionary Algorithm (GOMEA) for discrete optimization~\cite{bosman2012measures}, uses a tree-based representation of fixed size (i.e., a tree template) to enforce finding smaller expressions with high accuracy. In GOMEA, dependencies between genes are modelled, so that genes that have some dependency between them, are varied together. In GOMEA's GP variant, these genes are the template nodes. In each generation, this linkage between genes is modelled anew. GP-GOMEA has been found to be state-of-the-art with regards to its accuracy and model complexity trade-off on the Semantic Regression benchmark set, SRBench~\cite{la2021contemporary}.

\begin{figure}
     \centering
     \begin{subfigure}[b]{0.3\textwidth}
         \centering
         \includegraphics[width=\textwidth]{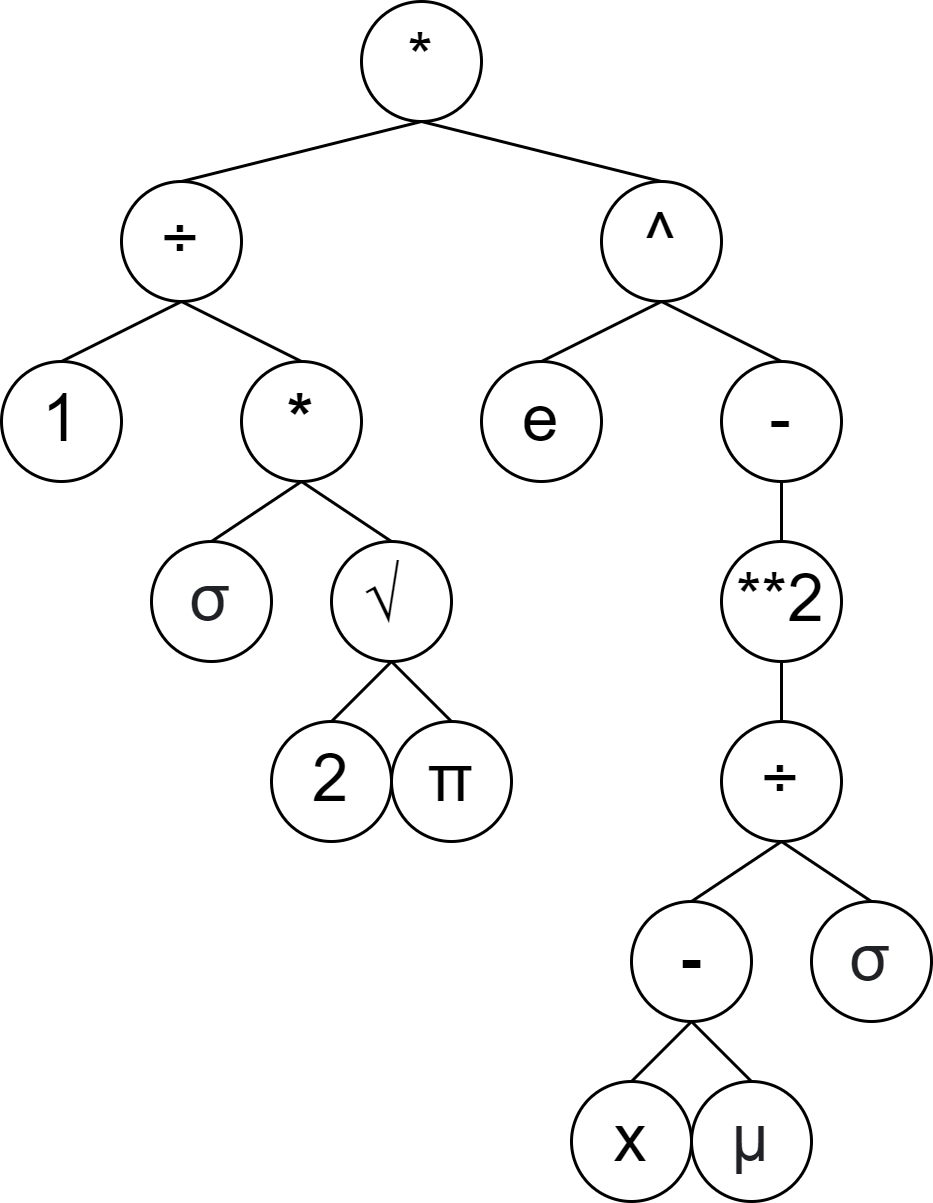}
         \caption{}
         \label{fig:example_large}
     \end{subfigure}
     \hfill
     \begin{subfigure}[b]{0.15\textwidth}
         \centering
         \includegraphics[width=\textwidth]{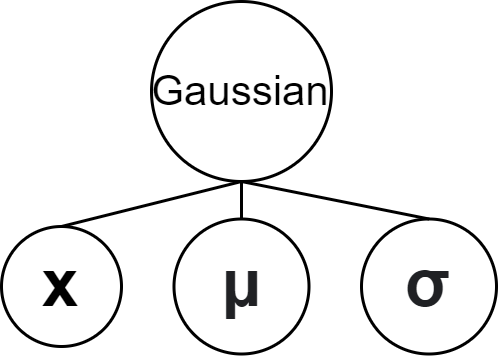}
         \caption{}
         \label{fig:example_small}
     \end{subfigure}
        \caption{Two trees, each describing the Gaussian probability density function. \ref{fig:example_small} makes use of a high-level operator.}
        \label{fig:guassian_example}
\end{figure}

In general, interpretability is promoted by shallow trees~\cite{mei2022explainable}. To this end, higher-level operators can also aid in abstracting subtrees into a more shallow version. Take, for instance, the Gaussian probability density function: with limited operators a tree modelling the Gaussian function can be seen in Figure \ref{fig:example_large}. However, when the Gaussian operator is available in the operator set, the large tree in Figure \ref{fig:example_large} can be abstracted into the tree shown in Figure \ref{fig:example_small}. With a mathematical background, the shallower tree is easier to interpret and preferred.

That being said, the use of higher-arity operators, such as the Gaussian function in the example above, has substantial impact on the efficiency and effectivity of GP-GOMEA. The reason for this is that the tree template needs to support the highest arity operator, creating very large templates of which potentially a large part is not used (if higher-arity operators are not frequently used). These unused parts are also known as \textit{introns}: nodes that are present in the template, but are not used. 
For instance, a parent node may represent an operator of arity 1 and, thus, may only require 1 child node. However, a fixed tree template with a branching factor of 2 will include a further (irrelevant) child by default. A tree may be shorter than the template, but the tree must still be padded with introns to fit the template. As the branching factor of a template increases, the number of potential introns increases exponentially.

Notwithstanding the current state-of-the-art performance of GP-GOMEA, we would still like to enhance its capabilities and performance with a focus on application in XAI. This includes the ability to abstract into shallower trees (for instance, like in Figure \ref{fig:guassian_example}). For this, we consider operators that are not only of higher arity but also take different data types as their arguments.

In particular, to model discontinuities in our target space we look to include higher-order cardinality operators, specifically: arithmetic Boolean logic as well as an \texttt{if-then-else} operator. These additions are aimed at modelling discontinuous relationships in data, whilst remaining interpretable. The \texttt{if-then-else} statement could then help express relationships dependent on certain conditions. For instance, when modelling developments in patients, a different model could be used dependent on a certain patient characteristic - simply expressed via an \texttt{if-then-else} statement in the final expression. 

We investigate the possibility to improve the search performed by GP-GOMEA to be better suited when dealing with higher-arity operators by threefold additions. First, we enlarge the operator set and include constraints that ensure correct expression types within operators. Secondly, we consider utilising the introns by optimising which children to use for an operator of lower arity than the template branching factor. Thirdly, when performing variation on an individual, we propose additionally inheriting entire subtrees from a different individual, provided both subtrees have the same parent operator. With this, we hope to share good building blocks that have evolved under a certain operator with individuals that have the same operator and increase the potential value of introns.

In summary, in this work we explore the following adaptations to GP-GOMEA: 
\begin{itemize}
    \item An enlarged operator set, that involves arithmetic, Boolean logic and ternary operators.    
    \item A semantic variation operation, that inherits subtrees from another solution in the population by considering subtrees located at a common parent operator.
    \item A child selection strategy that greedily selects the most suitable subtree (combination) for a parent within a current solution.
\end{itemize}
Finally, we investigate our adaptations through experiments on continuous and discontinuous expressions based on the Feynman equations.

\section{Background}
In this section, the background of the search enhancements made and their context in literature is given.
\paragraph{GP-GOMEA}
GOMEA was first introduced for optimisation with fixed-length binary strings as representation~\cite{bosman2012measures}. GOMEA was then adapted for GP by representing individuals with a tree template of a fixed size~\cite{virgolin2017scalable, virgolin2021improving}. The actual GP trees are built by sampling operators and terminals from their given respective sets as long as the trees fit within the template. The individuals, which are essentially still fixed-length strings, are then subject to the same notion of variation as in GOMEA. 

For variation, GP-GOMEA builds a linkage model in each generation. With this, dependencies are modelled between genes of a genotype, such that genes that are dependent on each other are varied together. By considering the linkage between genes in variation, GP-GOMEA aims to sustain and recombine good building blocks within individuals.

The dependency model in (GP-)GOMEA takes the form of a family of subsets (FOS), which consists of subsets of indices that indicate a position in the tree. To ensure that these positions are identified uniquely, they refer to indices in a tree's pre-order traversal~\cite{virgolin2017scalable}. Most often, the FOS is a linkage tree (LT), which is a hierarchical representation containing subsets of indices, i.e., for each subset $F_k$ in the LT with more than one gene, there also exists subsets $F_i$ and $F_j$, such that $F_i \cup F_j = F_k$ and $F_i \cap F_j = \emptyset$. Moreover, the set of all genes is included in the LT. To find which subsets of indices have some dependence between them and should thus be included in the FOS of a generation, pairwise mutual information is commonly used as a proxy. To build the LTs, a hierarchical agglomerative clustering procedure is used.

Once the FOS has been established, gene-pool optimal mixing (GOM) is performed on each solution in the population, which acts as a variation and selection procedure. The FOS is iterated through in a random order. For each element, a random donor individual from the population is selected. The values at the positions in the template indicated by the subset are then changed with the values at the same positions of the donor tree. If the result does not lead to a worse fitness, the change is accepted. Otherwise it is disregarded.

Our work presents two additions to this procedure, which are performed after the GOM phase. Specifically, we introduce semantic subtree inheritance and intron handling in the form of greedy child selection. While we describe these additions in section \ref{sec:methods}, here we describe their context and related literature.

\paragraph{Semantic Variation}
There has been substantial research into semantics in GP that has criticised many GP variants that solely consider syntax information in their genetic operators~\cite{vanneschi2014survey}.
The definition of the semantics of an individual refers to its output vector, given a certain input. This definition, for instance, gives way to variation operators that use semantic distances between two individuals to create an offspring, e.g., in \cite{nguyen2012investigation,beadle2008semantically}. For the semantic subtree inheritance in this work, we consider semantics within an individual.

\paragraph{Intron Handling}
In literature (e.g. in \cite{silva2009dynamic,carbajal2001evolutive}), introns refer to nodes whose execution do not affect a tree's output. Semantic introns are expressions that appear in the phenotype of an individual but do not influence the semantics of an individual. Semantic introns contribute to the bloating of individuals, which makes them less interpretable. In this work, we solely refer to syntactic introns: introns that do not appear in the phenotype of an individual and are only present in its genotype - specifically, in GP-GOMEA, this refers to completing a tree syntactically in a fixed-size tree template. 

Since the emergence of GP, there has been a debate on the impact of semantic introns. Some researchers believe introns protect building blocks of an individual, whereas others believe introns hamper the search~\cite{silva2009dynamic,carbajal2001evolutive}. As such, there have been different approaches to dealing with semantic introns. In variable length GP, the effect of adding explicitly defined introns, i.e., non-mutatable intron blocks whose purpose is to weight crossover behaviours~\cite{nordin1995explicitly,carbajal2001evolutive}, have been investigated. These explicitly defined introns were found to show slightly positive effects on the convergence behaviours~\cite{kouchakpour2009survey}.

A further approach to introns is their deletion after every fitness evaluation, although the effects of this are still inconclusive and the deletion produces overhead~\cite{castelli2013semantic}. To counteract bloating, GP-GOMEA limits the occurrence of semantic introns by limiting the solution size. This still allows for semantic introns but ensures that their number is limited.

In this work, the focus is henceforth only on syntactic introns. The handling of syntactic introns that appear in template-based GP has to the best of our knowledge not been extensively investigated.

\section{Methodology}
\label{sec:methods}
In this section, we elaborate on the details of our additions to GP-GOMEA. We first address the technical implications of modelling discontinuities, before both search extensions - semantic subtree inheritance and greedy child selection - are explained in depth.
\subsection{Modelling Clusters and Discontinuities}
Often it can be effective to decompose the complexity of a problem by splitting the problem into clusters. In each cluster, we can then describe different behaviours, which can lead to a more interpretable solution. Similarly, being able to model discontinuities directly in a regression could be easier to interpret than artificially approximating these.

To enhance the original GP-GOMEA to be able to model these clusters and discontinuities, arithmetic, Boolean logic, and the \texttt{if-then-else} operators are combined in an operator set. Boolean logic operators include but are not limited to, $>,<,==,$ \texttt{AND}, \texttt{OR}, and \texttt{NOT}.
Although this increases the branching factor of the tree from 2 to 3, the resulting piece-wise relationships are often more understandable. 

To ensure that the \texttt{if-then-else} operator, Boolean logic, and arithmetic operators are combined in a meaningful way, two syntactical constraints are imposed on the candidate solutions similar to how strongly typed GP works~\cite{montana1995strongly}. Both constraints are obeyed during population initialisation. During every change of an individual, it is checked whether a change has broken a constraint and, if so, the change is rejected.
The first constraint ensures that the first input to the \texttt{if-then-else} operator represents a Boolean. The second constraint specifies that an arithmetic operator requires numeric input. To protect interpretability, this means that Boolean outputs cannot be used in operators such as addition or multiplication.
As combining Boolean output via arithmetic operators allows to model discontinuities without using an \texttt{if-then-else} operator, this constraint can be deselected.  

\subsubsection{Ternary Operators}
\label{sec:Ternary Operators}
Since the inclusion of the \texttt{if-then-else} operator increases the branching factor of the tree template to 3, we also consider other operators that can be adapted to be of arity 3. Specifically, subtraction, addition, multiplication, as well as the Boolean logic operators \texttt{AND} and \texttt{OR} can all be used as operators of arity 2 as well as 3.

\subsection{Semantic Subtree Inheritance}
\begin{figure}
    \centering
    \includegraphics[width=0.4\textwidth]{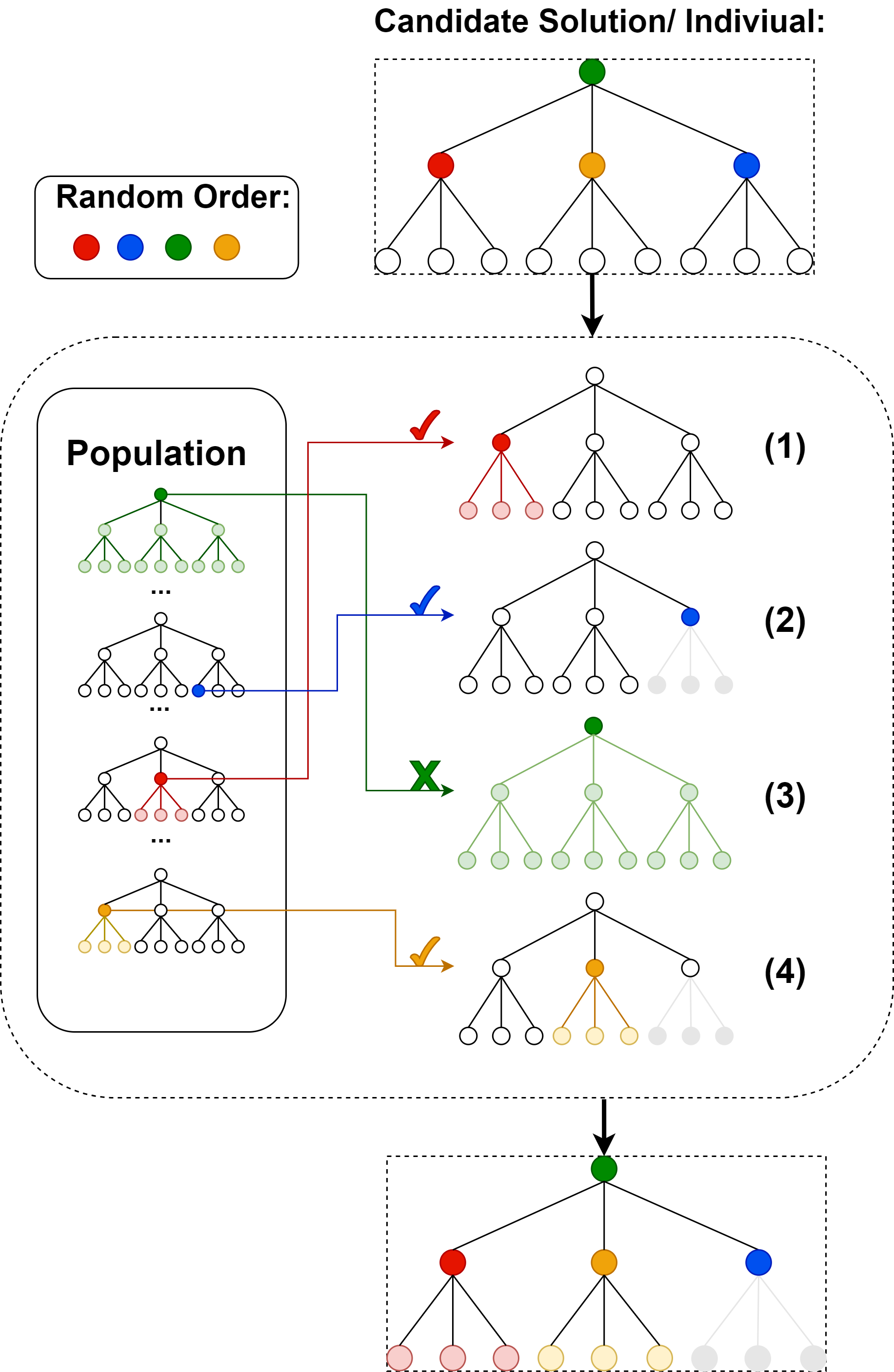}
    \caption{Semantic Subtree Inheritance: For each non-leaf, non-intron node, the node and its subtree are inherited based on its operator from a random donor individual from the population that includes the same operator somewhere in its tree. The ticks and crosses indicate that a change was accepted or rejected, respectively. The faint grey nodes indicate a subtree filled with syntactic introns.}
    \label{fig:SSS}
\end{figure}
During GOM, (combinations of) an individual's nodes, that have some linkage between them, are changed with the same (combinations of) nodes from another individual from the population. Hence, this inheritance operation solely considers the positions of nodes within an individual. We propose to not only consider the position of a node within a tree, but also to consider its semantic context within the individual. After the GOM procedure has been performed, \textit{semantic subtree inheritance} can be employed, as seen in Figure \ref{fig:SSS}. Here, we define subtrees with similar semantic contexts to be subtrees whose roots represent the same operator.

During the semantic subtree inheritance phase, all non-leaf, non-intron nodes of an individual are iterated through in random order. For each node, we search the population for a donor subtree that has at its root the same operator. We do this by randomly iterating through the population and stopping once a donor individual that uses the same operator somewhere is found. It is important to note that the position of the node within the tree is irrelevant, providing more flexibility to the variation operator in GP-GOMEA. Assuming a subtree has been found, the individual inherits the donor's subtree. Whenever this change results in better or equal fitness, the change is kept; it is otherwise reverted. No operation is performed in the case that no suitable donor is found. 

To ensure that a given donor subtree may fit into the fixed tree template of an individual, a semantically similar donor tree may only be selected given that said subtree is smaller or of equal depth to the current subtree. A subtree that is smaller than the current subtree can still be inherited. The missing nodes are then filled with syntactic introns - specifically, with the nodes that held those positions before the change.

\subsection{Greedy Child Selection}
\begin{figure}
    \centering
    \includegraphics[width=0.49\textwidth]{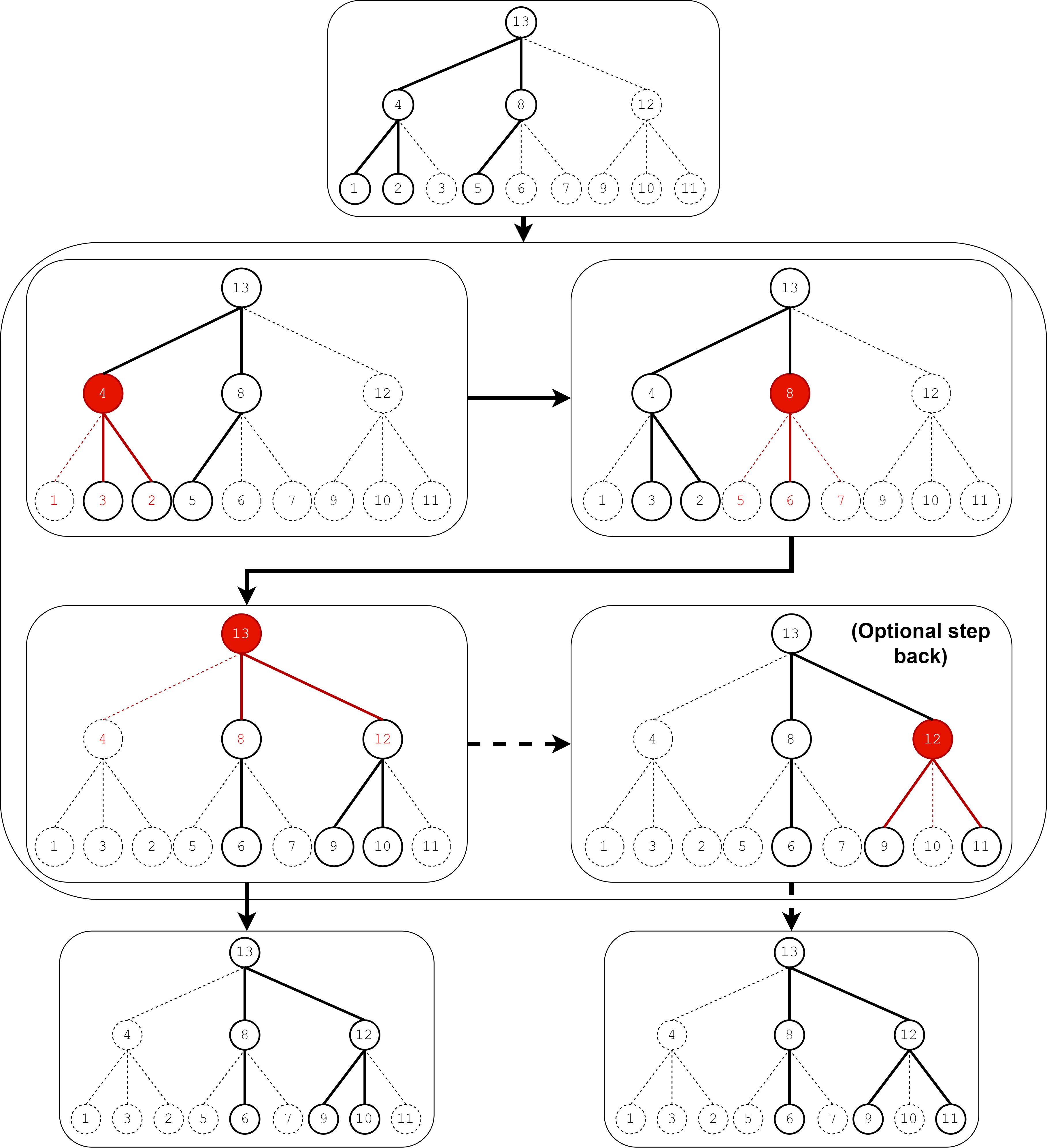}
    \caption{Greedy Child Selection: Nodes are visited in a depth-first traversal through the tree to select the child(ren) and their order with the best fit for the visited non-leaf, non-intron node. An optional step is to also visit a subtree that was previously unvisited as its root was an intron, which has now been selected to be used for its parent node.
    The figure shows this process for a tree with a branching factor of three. Thick lines indicate selected children, whereas dotted lines refer to intron children. The nodes show the following arities: 13, 4, and 12 are of arity 2, whereas 8 is of arity 1.}
    \label{fig:child_selection}
\end{figure}
The second search enhancement addresses the many syntactic introns that can occur in GP-GOMEA, especially once the operator set size and maximum operator arity are increased. Whenever the parent operator has an arity less than the template-arity, it will have at least one intron child. In the original GP-GOMEA, the left-most children are always selected. We propose introducing a greedy child selection strategy, as illustrated in Figure \ref{fig:child_selection}, to utilise the template's parts that typically stay unused, which is expected to increase especially when the template arity increases.

To utilise these introns, we propose to select the best (combination of) child(ren) for each parent in terms of fitness. Not only is it important to select which child(ren) to use, but for many operators, the order of the children plays a vital role and should be considered additionally. Further, it is important to note that each child selected impacts many other child selection decisions in the tree. Finding an optimal combination of the best (combination of) child(ren) for each parent in the tree simultaneously is an optimisation problem itself, for which exhaustive enumeration quickly becomes infeasible. Hence, we propose the selection in an iterative greedy manner. As the order in which nodes are visited is important, we iterate through the tree in a depth-first manner to visit every non-leaf, non-intron node and select its best children. 

Selecting the best child(ren) is done by iterating over all child selection options and recording their fitness. Finally, the child (combination) with the best fitness is accepted. The child selection at each node differs depending on a node's operator - specifically, an operator's arity and whether it is commutative. For instance, consider a ternary tree template, in which a node's children are referred to as \texttt{A}, \texttt{B}, and \texttt{C}. Here, a unary operator will have 3 options for its child (\texttt{A,B,C}), as does a commutative operator of arity 2 (\texttt{AB,AC,BC}). A binary operator which is not commutative has 6 combinations to choose from (\texttt{AB,AC,BA,BC,CA,CB}).
As previously mentioned in Section \ref{sec:Ternary Operators}, we have introduced the possibility for some binary operators to become ternary operators and select all three children. In the context of child selection, we, thus, also test whether a ternary operator would perform better than its binary version. As such, ternary operators that are commutative have one option (\texttt{ABC}), whereas non-commutative ternary operators have 6 options (\texttt{ABC,BCA,CAB,BAC,ACB,CBA}). 

Since the non-leaf, non-intron nodes are visited in a depth-first manner, a node's non-intron children (and their subtrees) have already been optimised. %visited
If an intron node is selected during this decision, its subtree has not yet been optimised and would typically not be visited in the context of a depth-first traversal. For this, an optional backtracking step is proposed, in which selected non-optimised introns are optimised before continuing the depth-first traversal to other nodes. 

As child selection uses function evaluations to assess the best child combination, there is a trade-off between child selection benefits and function evaluations used. Therefore, we consider greedy child selection in different configurations, specifically: chosen intron subtrees can optionally be added to the tree traversal. In addition to this, there is the option to only consider nodes of arity 1, up to arity 2, or to consider all arities, including their ternary variants. 

\section{Experimental Set-up}
First, we present the implementation used for GP-GOMEA in general, after which we elaborate on the various configurations of the introduced enhancements. We then introduce the symbolic regression problems, performance metrics, and experimental settings.
\subsection{GP-GOMEA Configuration}
In this section, the implementation and configurations of GP-GOMEA used for our experiments are given.
\subsubsection{Implementation of GP-GOMEA}
Since the introduction of GP-GOMEA~\cite{virgolin2017scalable, virgolin2021improving}, a new implementation has been made using the Eigen~\cite{eigenweb} package, as well as some additional minor adaptations, which are outlined in this section. 

The first adaptation refers to the usage of constants within linkage learning. For this, GP-GOMEA groups constants into bins, such that all constants within one bin are considered as a single symbol~\cite{virgolin2021improving}. To improve time efficiency, the new implementation places constants into 25 instead of 100 bins. Another adaptation addresses the situation that an individual is not improved after GOM (as well as semantic subtree inheritance and greedy child selection, if applicable). Previously, the individual would go through a new iteration of GOM, whereas the new implementation selects the individual into a tournament selection procedure of size 4, along with 3 random individuals from the population.

The new implementation uses an Interleaved Multistart Scheme (IMS) as proposed in \cite{virgolin2021improving}. IMS runs multiple evolutionary processes of increasing resources concurrently and advances these in an interleaved way. A further adaptation concerns the convergence criteria: for this, the population is sorted by fitness. A population is then considered converged once the fitness of 90\% of the least fit individuals is equal (given an error margin). This results in a convergence criterion that is met more quickly. Hence, populations that are not promising are aborted and restarted more often. Additionally, when restarting the population, the new implementation reinjects a randomly selected elite - this ensures that good building blocks already found are present in a new population.

\subsubsection{Search Extensions}
In total 14 configurations are tested: the original GP-GOMEA without search extensions, the greedy child selection (GCS) for (1) up to unary operators, (2) up to binary operators, and (3) all arities, including ternary variants. Further, the backtracking versions' of these three variants are considered, denoted by:  (1+),(2+), and (3+). The (3(+)) configurations, which consider ternary operators additionally, are only applicable when the template arity is bigger than 2. As such, we do not test these configurations on Binary templates. These 7 (for Binary trees, 5) configurations (the original GP-GOMEA and 6 variants of the greedy child selection, denoted as \textit{GCS}+variant ID) are tested with and without the semantic subtree inheritance (SSI) (denoted as \textit{SSI}), respectively. This gives way to 14 (for binary trees, 10) configurations. 

\subsection{Feynman Equations}
To evaluate the performances of different configurations in a controlled setting, ground truth symbolic regression problems are taken. We use the Feynman equations from SRBench~\cite{la2021contemporary} via the Penn Machine Learning Benchmark~\cite{romano2021pmlb}. To reduce overhead due to dataset size, the retrieved datasets were randomly downsampled from 100,000 data points to include 10,000 data points. For all Feynman equations, we calculated a feasible solution tree depth by hand. We use all Feynman equations that involve 4 variables and have feasible solutions at depths 3,4, and 5. This results in 22 problems: 9, 7, and 6 equations for depths, 3, 4, and 5 respectively. 

The selected Feynman equations are used as the benchmark set of continuous equations in our experiments. To create discontinuous problems, two Feynman equations are combined by adding a Boolean variable, which takes the value of 0 for the first equation and 1 for the second equation. Combining the equations per depth and disregarding the order gives $\frac{n(n-1)}{2}$ combinations, where $n$ is the number of datasets of a depth. This gives 36, 21, and 15 datasets for combinations of depth 3, 4, and 5 respectively. The Feynman equations and their discontinuous combinations give way to problems of varying difficulty (see supplementary material for more details). All datasets are randomly split into a training set of 75\% and a test set of 25\%.

\subsection{Performance Metrics}
To evaluate the performance of different configurations of GP-GOMEA with our enhancements, the R2 Error and the Mean Squared Error (MSE) are calculated. In all the experiments the MSE loss function is minimised.

To analyze the results statistically, we follow the recommendation of \cite{demvsar2006statistical}, which is implemented via Autorank~\cite{Herbold2020}. First, a Friedman test~\cite{friedman1940comparison} is performed to test whether the central tendencies in the population are equal to each other. We then employ the post hoc Nemenyi test~\cite{nemenyi1963distribution} for pairwise comparisons and calculate the critical distance, i.e., the minimum significant difference.
For further visualisation, we take inspiration from \cite{la2021contemporary}.

\subsection{Experiments}
It is important to note that in all our experiments, we ensure that each configuration is tested with the same amount of resources. Since the enhancements require more evaluations within one generation, the resources are measured in terms of function evaluations, instead of generations. Additionally, we track the performances of each configuration at multiple checkpoints, to investigate how the configurations impact the optimisation process over time. Specifically, we record the performance at checkpoints of 100, 500, 1,000, 5,000, 10,000, 50,000, 100,000, 500,000, 1,000,000, and 5,000,000 evaluations. As GP is a stochastic process, each experiment is repeated 20 times. For analysis, the median performance metrics across random seeds are considered. Further, we avoid setting the population size manually by employing IMS as proposed in~\cite{virgolin2021improving} with a starting population of 64 individuals and 10 intermediate generations between evolutions. 

\subsubsection{Operators}
\begin{table}[]
    \centering
    \tabcolsep=1.25mm
    \begin{tabular}{|c|c|p{4.5cm}|}
         \hline
         \scalebox{0.9}{Branch factor} & \scalebox{0.9}{\# Operators}& \scalebox{0.9}{Operator Set} \\
         \hline
         3 & 22 &\tiny $+, -x, -, *, \frac{1}{x}, /, sin, cos, log, exp, sqrt, x^2, x^3, x^4,\linebreak x^5, =, >, < \texttt{IfThenElse}, \texttt{AND}, \texttt{OR}, \texttt{NOT}$\\
          3 & 11 &\tiny $+, -, *, /, sin, log, sqrt, x^2, x^3, <, \texttt{IfThenElse}$\\
          \hline
         2 & 15 &\tiny $+, -x, -, *, \frac{1}{x}, /, sin, cos, log, exp, sqrt, x^2, x^3, x^4, x^5$ \\         
         2 & 9 &\tiny $+, -, *, /, sin, log, sqrt, x^2, x^3$\\
         2 & 4 &\tiny $+, -, *, /$\\
         \hline
    \end{tabular}
    \caption{The operator sets and their branching factor used in the experiments: ordered by the complexity of their induced search space (from most to least complex). The maximum arity of the operator sets induces the template arity. The set of 22 operators and 15 operators include the most common operators in the Feynman equations and all operators needed to solve the selected problems. $-x$ denotes the unary minus operator. }
    \label{tab:ops}
\end{table}
The enhancements introduced in this work aim to improve search efficiency, especially when the operator set and the template arity are increased. To investigate what effect the different GP-GOMEA enhancements have for different search spaces, we test different sets of operators (see Table \ref{tab:ops}). With the selected operator sets, we vary the complexity of the search space in terms of the operator set size and tree template size.

 Depending on the operator set, different enhancement configurations may not be applicable. Due to this, the greedy child selection which considers ternary operators is not applied on the operator sets with a branching factor of 2. Further, as the operator set with 4 operators only includes binary operators, the greedy child selection configuration considering only unary operators is the same as not applying greedy child selection.
\subsubsection{Depth}
The tree template does not only grow with operators of higher arity but also grows with increasing depth. Using the calculated feasible depths for each problem respectively, we perform experiments across different depths. 
In addition to this, we investigate the behaviour of the enhancements when the depth is smaller than required. To investigate this, the maximum depth was set to the calculated feasible solution depth minus 1. 

\section{Results \& Discussion}
\begin{figure}
    \centering
    \includegraphics[width = 0.5\textwidth]{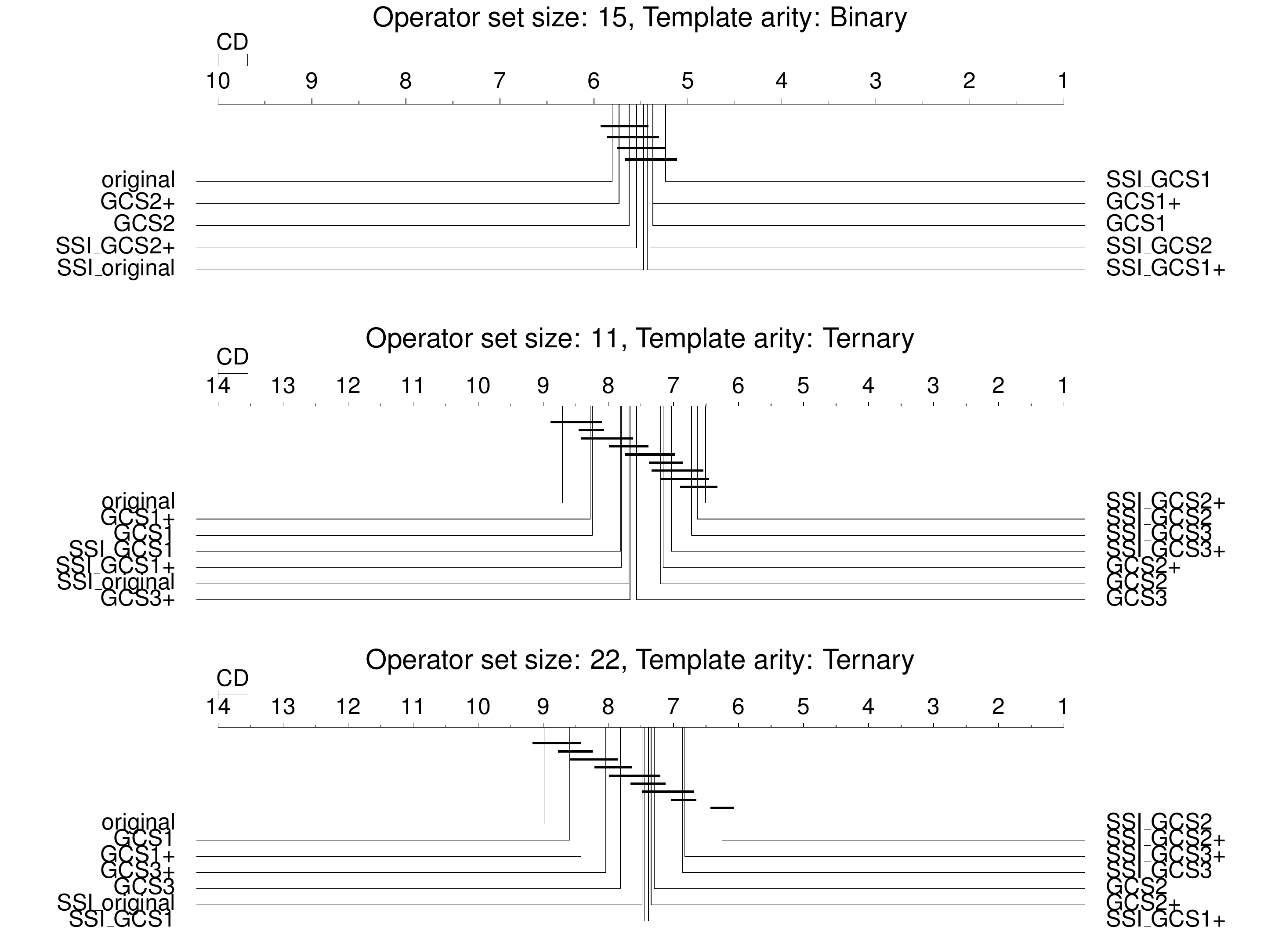}
    \caption{Statistical significant differences between the mean rank of GP-GOMEA configurations. Two configurations that are connected by a bar are not significantly different, whereas configurations that are further apart than the critical distance (CD) are statistically significantly different.}%The critical distance that must be between two mean ranks to be significantly different is }
    \label{fig:stats}
\end{figure}
\begin{figure*}
    \centering
    \includegraphics[width =\textwidth]{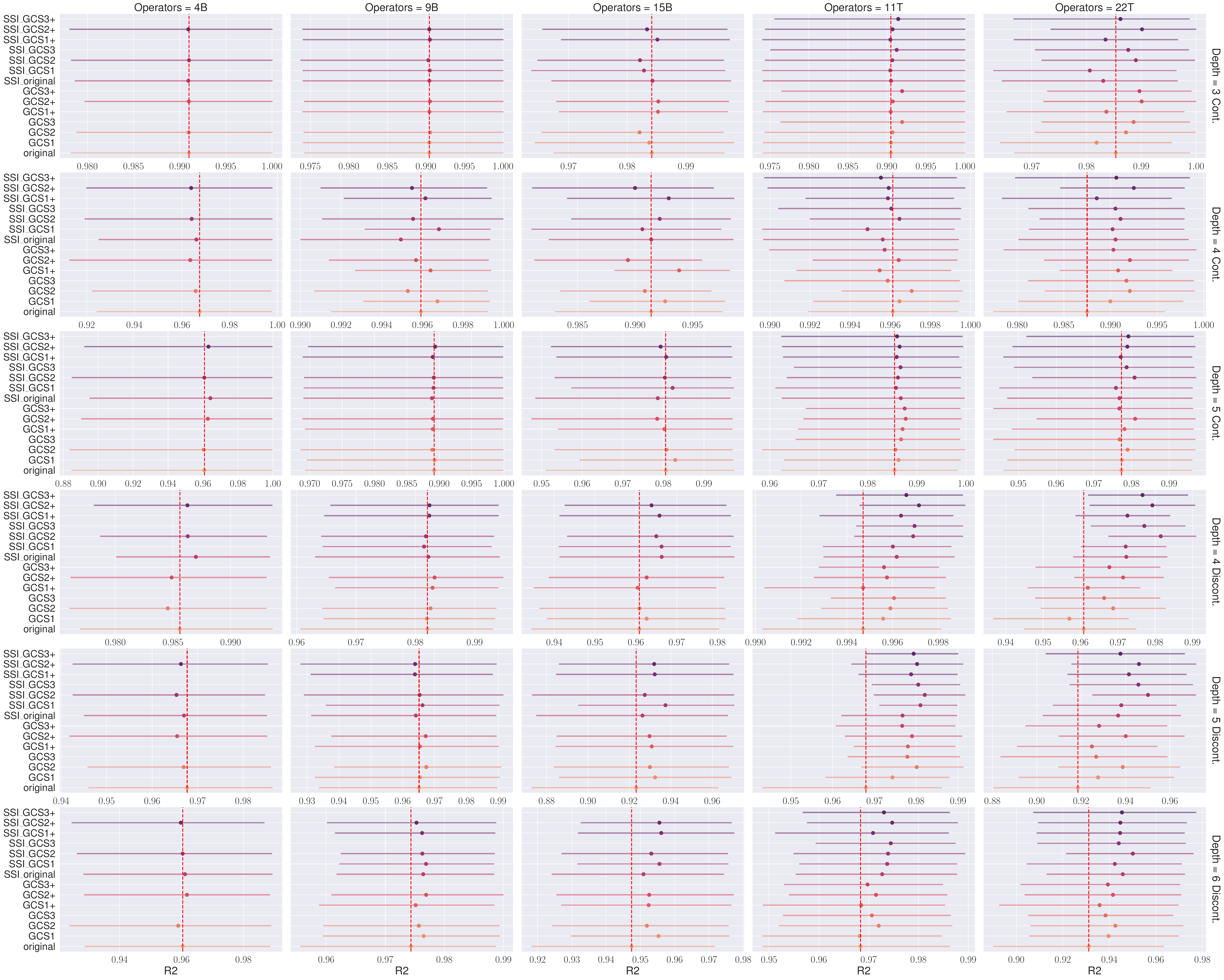}
    \caption{The impact of varying the operators on the R2 metrics of different GP-GOMEA configurations on continuous and discontinuous problems of different depths. The performances are measured after 5,000,000 function evaluations. A B indicates a binary tree, whereas a T indicates a ternary tree. Points indicate the mean of the median R2 performances per dataset, whereas the bars indicate the 95\% confidence interval. To ease comparison, the red vertical line shows the mean performance of the original GP-GOMEA configuration in the respective setting.}
    \label{fig:all}
\end{figure*}
\begin{figure}
    \centering
    \includegraphics[width=1.1\linewidth]{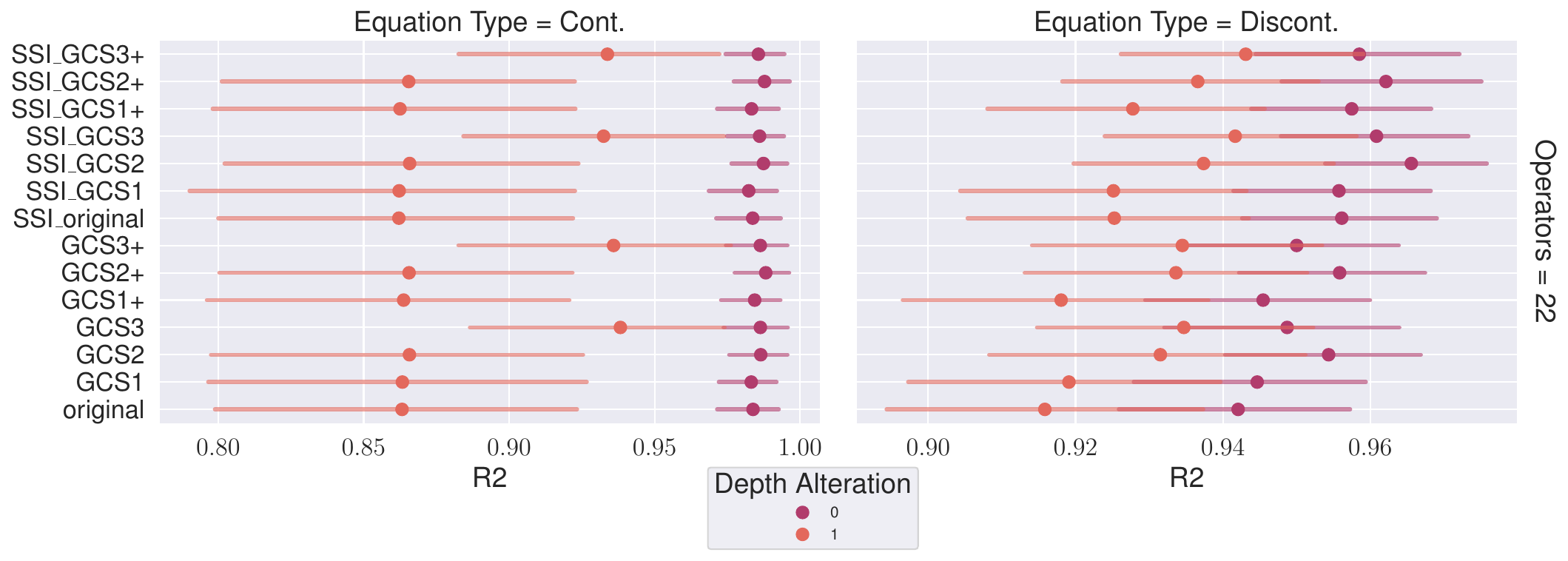}
    \caption{R2 metrics of different GP-GOMEA configurations on continuous and discontinuous problems with 22 operators. The colours show whether a smaller tree depth was used. Points indicate the mean of the median R2 performance on all continuous and discontinuous problems, whereas the bars indicate the 95\% confidence interval.}
    \label{fig:limDepth}
\end{figure}
This section presents the results of the experiments performed and outlines insights that stem from them.
\subsection{Performance across Search Complexities}

Figure \ref{fig:all} summarises the performance of the different GP-GOMEA configurations after 5,000,000 function evaluations using the different operator sets seen in Table ~\ref{tab:ops}. Further, the performance of each configuration is shown across discontinuous and continuous Feynman equations of different depths.

In general, Figure \ref{fig:all} shows a positive effect of the different enhancement configurations to the original GP-GOMEA once the search space becomes more complex. The benefit with added complexity can be seen with regards to larger template arities and larger operator sets (the rightmost columns), as well as with more complex problems that are deeper and/or discontinuous (the lower rows). In other situations, the search enhancements do not worsen the performance of the original GP-GOMEA.

Since all configurations across all experiments were given the same amount of resources, the configurations in different search spaces are stopped in different phases of their evolutionary process. 

This means that for search spaces with fewer available operators and a smaller tree template, all variants have achieved more or less the best they can. For the larger search spaces, we see differences because some configurations are more efficient than others. In particular, the proposed enhancements have a positive effect as can be seen from Figure \ref{fig:stats} and \ref{fig:all}. Given sufficient resources, we expect that the performances will be similar across configurations (comparable to the less complex search spaces in Figure \ref{fig:all}). Conversely, for these less complex spaces, we expect to see differences (albeit potentially smaller) earlier during the search.

We briefly investigate this in the supplementary material and find that the enhancements improve the performance at fewer function evaluations when applied with fewer operators and a smaller template arity. Specifically, across all problems, the enhancements are significantly better than the original GP-GOMEA after 500 and 1,000 function evaluations, which supports our hypothesis. Further, we also observe an interplay between IMS and configurations.

\subsubsection{Statistical Analysis}
A statistical analysis verified the above-mentioned notion that (after 5,000,000 function evaluations) configurations using the search enhancements perform superior to the original GP-GOMEA configuration, especially when the operator set size is large and/or higher-order cardinality functions are used.
The differences in performance are not statistically significant between the configurations of GP-GOMEA for the operator sets of size 4 and 9. In Figure \ref{fig:stats}, we show the statistical differences found for the operator sets of size 11, 15, and 22. It can be seen that with a maximum arity of three the enhancements are more useful than for a smaller template. That being said, the benefit of the enhancements also increases with the number of operators used.

From this, the configurations including greedy child selection that considers operators up to an arity of 2 are most successful in ternary trees, whereas for binary trees the greedy child selection considering operators of arity of 1 are beneficial. In all cases, semantic subtree inheritance seems beneficial.

\subsection{Performance across forced smaller solution sizes}
In this section, we analyse the effect that a smaller solution size can have on the different configurations. For this, results are obtained using an operator set size of 22 with the correct depth (presented in the previous section) and compared with the configurations using a depth that is smaller by 1.

Figure \ref{fig:limDepth} shows the performance of various configurations when given less depth than the feasible solution size for each problem. For all problems, one can see that the performance decreases when using a smaller solution size than needed. For discontinuous problems, the performances of both the correct depth and the altered depth lie closer to each other. For the altered depth, the configurations that consider ternary operators, as well as use semantic subtree inheriting perform the best. 

For continuous problems, the benefit of considering ternary operators is highlighted, as all configurations that do so (regardless of additional backtracking or semantic subtree inheritance) significantly outperform the configurations that do not. This is intuitive as configurations that can flexibly adapt the arity of some binary operators can model a problem in a smaller solution size. The addition of ternary operators could become even more relevant when the depth is unknown (as is common) or pressure is added to reduce complexity in GP-GOMEA (e.g., through a second objective). From an XAI perspective being able to increase performance whilst retaining a less deep solution template is likely to aid interpretability.

\section{Limitations and Future Work}
The semantic subtree inheritance could be adapted further, such that partial subtrees, i.e., single arguments to a parent operator, could be inherited as well as entire subtrees. For further flexibility, the leaf nodes could also be considered during semantic subtree inheritance, such that variables and groups of constants could be inherited. This would likely reduce the required population size because required constant and/or input variable index values do not need to exist with nonzero probability in each position upon initialisation anymore (which is currently the case due to the variation performed in the original GP-GOMEA).

Further, an explicit comparison between the configurations of GP-GOMEA and traditional constrained (w.r.t. height) GP should be made, specifically with higher-order cardinalities. A previous comparison of binary templates between GP-GOMEA and constrained GP found GP-GOMEA was more effective than GP~\cite{virgolin2021improving}. Based on this, it can be expected that this tendency will hold for higher cardinalities - nevertheless, this should be investigated.

Moreover, future work could test the different configurations of GP-GOMEA with larger operator sets, including multiple operators of higher arity, on a varying set of benchmark data, such as SRBench~\cite{la2021contemporary} as well as real-world data. Designing additional discontinuous benchmark problems would also be an important avenue of future research that could benefit the GP community. 

\section{Conclusion}
In GP, high-level operators of higher cardinality can aid in creating shallower, more interpretable solution trees. In our work, we aim to improve the search efficiency of GP-GOMEA, such that GP-GOMEA can handle a more complex search space in terms of larger tree templates and larger operator sets more efficiently. 

For this, we have proposed two novel enhancements to the search in GP-GOMEA: semantic subtree inheritance and greedy child selection. Semantic subtree inheritance performs additional variation by inheriting subtrees that have a common parent operator. Further, the greedy child selection utilises the potentially increasing number of syntactic introns present with a larger tree template by greedily selecting the best (combination) of child(ren).

We performed an in-depth experimental investigation of the two proposed search enhancements on various tree templates, operator sets, and depths. For this, a selection of the ground truth Feynmann equations are used, which were extended to create discontinuous problems. 

We show that our proposed enhancements improve search efficiency in many cases, and do not decrease the search efficiency in all cases. The enhancements were shown to statistically improve search, especially when increasing the operator set size and/or including higher-order cardinality functions. 

We believe that larger operator sets and/or more complex operators can allow GP-GOMEA to model complex functions more interpretably. Hence, increasing the search efficiency for GP-GOMEA to handle this more complex search space is an important step for its use in XAI.

\vspace*{-2mm}
\begin{acks}
This research is part of the "Uitlegbare Kunstmatige Intelligentie" project funded by the Stichting Gieskes-Strijbis Fonds. Furthermore, we thank NWO for the Small Compute grant on the Dutch National Supercomputer Snellius.
\end{acks}

\bibliographystyle{ACM-Reference-Format}
\bibliography{actual_bib}


\begin{thebibliography}{24}

%%% the Web address.

\ifx \showCODEN    \undefined \def \showCODEN     #1{\unskip}     \fi
\ifx \showDOI      \undefined \def \showDOI       #1{#1}\fi
\ifx \showISBNx    \undefined \def \showISBNx     #1{\unskip}     \fi
\ifx \showISBNxiii \undefined \def \showISBNxiii  #1{\unskip}     \fi
\ifx \showISSN     \undefined \def \showISSN      #1{\unskip}     \fi
\ifx \showLCCN     \undefined \def \showLCCN      #1{\unskip}     \fi
\ifx \shownote     \undefined \def \shownote      #1{#1}          \fi
\ifx \showarticletitle \undefined \def \showarticletitle #1{#1}   \fi
\ifx \showURL      \undefined \def \showURL       {\relax}        \fi
% invisible to TeX
\providecommand\bibfield[2]{#2}
\providecommand\bibinfo[2]{#2}
\providecommand\natexlab[1]{#1}
\providecommand\showeprint[2][]{arXiv:#2}

\bibitem[\protect\citeauthoryear{Beadle and Johnson}{Beadle and Johnson}{2008}]%
        {beadle2008semantically}
\bibfield{author}{\bibinfo{person}{Lawrence Beadle} {and} \bibinfo{person}{Colin~G Johnson}.} \bibinfo{year}{2008}\natexlab{}.
\newblock \showarticletitle{Semantically driven crossover in genetic programming}. In \bibinfo{booktitle}{\emph{2008 IEEE Congress on Evolutionary Computation (IEEE World Congress on Computational Intelligence)}}. IEEE, \bibinfo{pages}{111--116}.
\newblock


\bibitem[\protect\citeauthoryear{Bosman and Thierens}{Bosman and Thierens}{2012}]%
        {bosman2012measures}
\bibfield{author}{\bibinfo{person}{Peter~AN Bosman} {and} \bibinfo{person}{Dirk Thierens}.} \bibinfo{year}{2012}\natexlab{}.
\newblock \showarticletitle{On measures to build linkage trees in LTGA}. In \bibinfo{booktitle}{\emph{International Conference on Parallel Problem Solving from Nature}}. Springer, \bibinfo{pages}{276--285}.
\newblock


\bibitem[\protect\citeauthoryear{Carbajal and Martinez}{Carbajal and Martinez}{2001}]%
        {carbajal2001evolutive}
\bibfield{author}{\bibinfo{person}{Santiago Garci{\'{}}~a Carbajal} {and} \bibinfo{person}{Ferm{\'{}} in~Gonz{\'a}lez Martinez}.} \bibinfo{year}{2001}\natexlab{}.
\newblock \showarticletitle{Evolutive introns: A non-costly method of using introns in GP}.
\newblock \bibinfo{journal}{\emph{Genetic Programming and Evolvable Machines}}  \bibinfo{volume}{2} (\bibinfo{year}{2001}), \bibinfo{pages}{111--122}.
\newblock


\bibitem[\protect\citeauthoryear{Castelli, Vanneschi, and Silva}{Castelli et~al\mbox{.}}{2013}]%
        {castelli2013semantic}
\bibfield{author}{\bibinfo{person}{Mauro Castelli}, \bibinfo{person}{Leonardo Vanneschi}, {and} \bibinfo{person}{Sara Silva}.} \bibinfo{year}{2013}\natexlab{}.
\newblock \showarticletitle{Semantic search-based genetic programming and the effect of intron deletion}.
\newblock \bibinfo{journal}{\emph{IEEE Transactions on Cybernetics}} \bibinfo{volume}{44}, \bibinfo{number}{1} (\bibinfo{year}{2013}), \bibinfo{pages}{103--113}.
\newblock


\bibitem[\protect\citeauthoryear{Dem{\v{s}}ar}{Dem{\v{s}}ar}{2006}]%
        {demvsar2006statistical}
\bibfield{author}{\bibinfo{person}{Janez Dem{\v{s}}ar}.} \bibinfo{year}{2006}\natexlab{}.
\newblock \showarticletitle{Statistical comparisons of classifiers over multiple data sets}.
\newblock \bibinfo{journal}{\emph{The Journal of Machine learning research}}  \bibinfo{volume}{7} (\bibinfo{year}{2006}), \bibinfo{pages}{1--30}.
\newblock


\bibitem[\protect\citeauthoryear{Friedman}{Friedman}{1940}]%
        {friedman1940comparison}
\bibfield{author}{\bibinfo{person}{Milton Friedman}.} \bibinfo{year}{1940}\natexlab{}.
\newblock \showarticletitle{A comparison of alternative tests of significance for the problem of m rankings}.
\newblock \bibinfo{journal}{\emph{The Annals of Mathematical Statistics}} \bibinfo{volume}{11}, \bibinfo{number}{1} (\bibinfo{year}{1940}), \bibinfo{pages}{86--92}.
\newblock


\bibitem[\protect\citeauthoryear{Guennebaud, Jacob, et~al\mbox{.}}{Guennebaud et~al\mbox{.}}{2010}]%
        {eigenweb}
\bibfield{author}{\bibinfo{person}{Ga\"{e}l Guennebaud}, \bibinfo{person}{Beno\^{i}t Jacob}, {et~al\mbox{.}}} \bibinfo{year}{2010}\natexlab{}.
\newblock \bibinfo{title}{Eigen v3}.
\newblock \bibinfo{howpublished}{http://eigen.tuxfamily.org}.
\newblock


\bibitem[\protect\citeauthoryear{Harrison, Alderliesten, and Bosman}{Harrison et~al\mbox{.}}{2022}]%
        {harrison2022gene}
\bibfield{author}{\bibinfo{person}{Joe Harrison}, \bibinfo{person}{Tanja Alderliesten}, {and} \bibinfo{person}{Peter~AN Bosman}.} \bibinfo{year}{2022}\natexlab{}.
\newblock \showarticletitle{Gene-pool Optimal Mixing in Cartesian Genetic Programming}. In \bibinfo{booktitle}{\emph{International Conference on Parallel Problem Solving from Nature}}. Springer, \bibinfo{pages}{19--32}.
\newblock


\bibitem[\protect\citeauthoryear{Herbold}{Herbold}{2020}]%
        {Herbold2020}
\bibfield{author}{\bibinfo{person}{Steffen Herbold}.} \bibinfo{year}{2020}\natexlab{}.
\newblock \showarticletitle{Autorank: A Python package for automated ranking of classifiers}.
\newblock \bibinfo{journal}{\emph{Journal of Open Source Software}} \bibinfo{volume}{5}, \bibinfo{number}{48} (\bibinfo{year}{2020}), \bibinfo{pages}{2173}.
\newblock
\urldef\tempurl%
\url{https://doi.org/10.21105/joss.02173}
\showDOI{\tempurl}


\bibitem[\protect\citeauthoryear{Kouchakpour, Zaknich, and Braeunl}{Kouchakpour et~al\mbox{.}}{2009}]%
        {kouchakpour2009survey}
\bibfield{author}{\bibinfo{person}{Peyman Kouchakpour}, \bibinfo{person}{Anthony Zaknich}, {and} \bibinfo{person}{Thomas Braeunl}.} \bibinfo{year}{2009}\natexlab{}.
\newblock \showarticletitle{A survey and taxonomy of performance improvement of canonical genetic programming}.
\newblock \bibinfo{journal}{\emph{Knowledge and information systems}}  \bibinfo{volume}{21} (\bibinfo{year}{2009}), \bibinfo{pages}{1--39}.
\newblock


\bibitem[\protect\citeauthoryear{La~Cava, Orzechowski, Burlacu, de~Fran{\c{c}}a, Virgolin, Jin, Kommenda, and Moore}{La~Cava et~al\mbox{.}}{2021}]%
        {la2021contemporary}
\bibfield{author}{\bibinfo{person}{William La~Cava}, \bibinfo{person}{Patryk Orzechowski}, \bibinfo{person}{Bogdan Burlacu}, \bibinfo{person}{Fabr{\'\i}cio~Olivetti de Fran{\c{c}}a}, \bibinfo{person}{Marco Virgolin}, \bibinfo{person}{Ying Jin}, \bibinfo{person}{Michael Kommenda}, {and} \bibinfo{person}{Jason~H Moore}.} \bibinfo{year}{2021}\natexlab{}.
\newblock \showarticletitle{Contemporary symbolic regression methods and their relative performance}.
\newblock \bibinfo{journal}{\emph{arXiv preprint arXiv:2107.14351}} (\bibinfo{year}{2021}).
\newblock


\bibitem[\protect\citeauthoryear{Lipton}{Lipton}{2018}]%
        {lipton2018mythos}
\bibfield{author}{\bibinfo{person}{Zachary~C Lipton}.} \bibinfo{year}{2018}\natexlab{}.
\newblock \showarticletitle{The mythos of model interpretability: In machine learning, the concept of interpretability is both important and slippery.}
\newblock \bibinfo{journal}{\emph{Queue}} \bibinfo{volume}{16}, \bibinfo{number}{3} (\bibinfo{year}{2018}), \bibinfo{pages}{31--57}.
\newblock


\bibitem[\protect\citeauthoryear{Mei, Chen, Lensen, Xue, and Zhang}{Mei et~al\mbox{.}}{2022}]%
        {mei2022explainable}
\bibfield{author}{\bibinfo{person}{Yi Mei}, \bibinfo{person}{Qi Chen}, \bibinfo{person}{Andrew Lensen}, \bibinfo{person}{Bing Xue}, {and} \bibinfo{person}{Mengjie Zhang}.} \bibinfo{year}{2022}\natexlab{}.
\newblock \showarticletitle{Explainable artificial intelligence by genetic programming: A survey}.
\newblock \bibinfo{journal}{\emph{IEEE Transactions on Evolutionary Computation}} (\bibinfo{year}{2022}).
\newblock


\bibitem[\protect\citeauthoryear{Montana}{Montana}{1995}]%
        {montana1995strongly}
\bibfield{author}{\bibinfo{person}{David~J Montana}.} \bibinfo{year}{1995}\natexlab{}.
\newblock \showarticletitle{Strongly typed genetic programming}.
\newblock \bibinfo{journal}{\emph{Evolutionary computation}} \bibinfo{volume}{3}, \bibinfo{number}{2} (\bibinfo{year}{1995}), \bibinfo{pages}{199--230}.
\newblock


\bibitem[\protect\citeauthoryear{Nemenyi}{Nemenyi}{1963}]%
        {nemenyi1963distribution}
\bibfield{author}{\bibinfo{person}{Peter~Bjorn Nemenyi}.} \bibinfo{year}{1963}\natexlab{}.
\newblock \bibinfo{booktitle}{\emph{Distribution-free multiple comparisons.}}
\newblock \bibinfo{publisher}{Princeton University}.
\newblock


\bibitem[\protect\citeauthoryear{Nguyen, Nguyen, O’Neill, and Agapitos}{Nguyen et~al\mbox{.}}{2012}]%
        {nguyen2012investigation}
\bibfield{author}{\bibinfo{person}{Quang~Uy Nguyen}, \bibinfo{person}{Xuan~Hoai Nguyen}, \bibinfo{person}{Michael O’Neill}, {and} \bibinfo{person}{Alexandros Agapitos}.} \bibinfo{year}{2012}\natexlab{}.
\newblock \showarticletitle{An investigation of fitness sharing with semantic and syntactic distance metrics}. In \bibinfo{booktitle}{\emph{Genetic Programming: 15th European Conference, EuroGP 2012, M{\'a}laga, Spain, April 11-13, 2012. Proceedings 15}}. Springer, \bibinfo{pages}{109--120}.
\newblock


\bibitem[\protect\citeauthoryear{Nordin, Francone, and Banzhaf}{Nordin et~al\mbox{.}}{1995}]%
        {nordin1995explicitly}
\bibfield{author}{\bibinfo{person}{Peter Nordin}, \bibinfo{person}{Frank Francone}, {and} \bibinfo{person}{Wolfgang Banzhaf}.} \bibinfo{year}{1995}\natexlab{}.
\newblock \showarticletitle{Explicitly de ned introns and destructive crossover in genetic programming}.
\newblock \bibinfo{journal}{\emph{Advances in genetic programming}}  \bibinfo{volume}{2} (\bibinfo{year}{1995}), \bibinfo{pages}{111--134}.
\newblock


\bibitem[\protect\citeauthoryear{Romano, Le, La~Cava, Gregg, Goldberg, Chakraborty, Ray, Himmelstein, Fu, and Moore}{Romano et~al\mbox{.}}{2021}]%
        {romano2021pmlb}
\bibfield{author}{\bibinfo{person}{Joseph~D Romano}, \bibinfo{person}{Trang~T Le}, \bibinfo{person}{William La~Cava}, \bibinfo{person}{John~T Gregg}, \bibinfo{person}{Daniel~J Goldberg}, \bibinfo{person}{Praneel Chakraborty}, \bibinfo{person}{Natasha~L Ray}, \bibinfo{person}{Daniel Himmelstein}, \bibinfo{person}{Weixuan Fu}, {and} \bibinfo{person}{Jason~H Moore}.} \bibinfo{year}{2021}\natexlab{}.
\newblock \showarticletitle{PMLB v1.0: an open source dataset collection for benchmarking machine learning methods}.
\newblock \bibinfo{journal}{\emph{arXiv preprint arXiv:2012.00058v2}} (\bibinfo{year}{2021}).
\newblock


\bibitem[\protect\citeauthoryear{Sijben, Alderliesten, and Bosman}{Sijben et~al\mbox{.}}{2022}]%
        {sijben2022multi}
\bibfield{author}{\bibinfo{person}{Evi Sijben}, \bibinfo{person}{Tanja Alderliesten}, {and} \bibinfo{person}{Peter~AN Bosman}.} \bibinfo{year}{2022}\natexlab{}.
\newblock \showarticletitle{Multi-modal multi-objective model-based genetic programming to find multiple diverse high-quality models}. In \bibinfo{booktitle}{\emph{Proceedings of the Genetic and Evolutionary Computation Conference}}. \bibinfo{pages}{440--448}.
\newblock


\bibitem[\protect\citeauthoryear{Silva and Costa}{Silva and Costa}{2009}]%
        {silva2009dynamic}
\bibfield{author}{\bibinfo{person}{Sara Silva} {and} \bibinfo{person}{Ernesto Costa}.} \bibinfo{year}{2009}\natexlab{}.
\newblock \showarticletitle{Dynamic limits for bloat control in genetic programming and a review of past and current bloat theories}.
\newblock \bibinfo{journal}{\emph{Genetic Programming and Evolvable Machines}}  \bibinfo{volume}{10} (\bibinfo{year}{2009}), \bibinfo{pages}{141--179}.
\newblock


\bibitem[\protect\citeauthoryear{Vanneschi, Castelli, and Silva}{Vanneschi et~al\mbox{.}}{2014}]%
        {vanneschi2014survey}
\bibfield{author}{\bibinfo{person}{Leonardo Vanneschi}, \bibinfo{person}{Mauro Castelli}, {and} \bibinfo{person}{Sara Silva}.} \bibinfo{year}{2014}\natexlab{}.
\newblock \showarticletitle{A survey of semantic methods in genetic programming}.
\newblock \bibinfo{journal}{\emph{Genetic Programming and Evolvable Machines}}  \bibinfo{volume}{15} (\bibinfo{year}{2014}), \bibinfo{pages}{195--214}.
\newblock


\bibitem[\protect\citeauthoryear{Vilone and Longo}{Vilone and Longo}{2020}]%
        {vilone2020explainable}
\bibfield{author}{\bibinfo{person}{Giulia Vilone} {and} \bibinfo{person}{Luca Longo}.} \bibinfo{year}{2020}\natexlab{}.
\newblock \showarticletitle{Explainable artificial intelligence: a systematic review}.
\newblock \bibinfo{journal}{\emph{arXiv preprint arXiv:2006.00093}} (\bibinfo{year}{2020}).
\newblock


\bibitem[\protect\citeauthoryear{Virgolin, Alderliesten, Witteveen, and Bosman}{Virgolin et~al\mbox{.}}{2017}]%
        {virgolin2017scalable}
\bibfield{author}{\bibinfo{person}{Marco Virgolin}, \bibinfo{person}{Tanja Alderliesten}, \bibinfo{person}{Cees Witteveen}, {and} \bibinfo{person}{Peter~AN Bosman}.} \bibinfo{year}{2017}\natexlab{}.
\newblock \showarticletitle{Scalable genetic programming by gene-pool optimal mixing and input-space entropy-based building-block learning}. In \bibinfo{booktitle}{\emph{Proceedings of the Genetic and Evolutionary Computation Conference}}. \bibinfo{pages}{1041--1048}.
\newblock


\bibitem[\protect\citeauthoryear{Virgolin, Alderliesten, Witteveen, and Bosman}{Virgolin et~al\mbox{.}}{2021}]%
        {virgolin2021improving}
\bibfield{author}{\bibinfo{person}{Marco Virgolin}, \bibinfo{person}{Tanja Alderliesten}, \bibinfo{person}{Cees Witteveen}, {and} \bibinfo{person}{Peter~AN Bosman}.} \bibinfo{year}{2021}\natexlab{}.
\newblock \showarticletitle{Improving model-based genetic programming for symbolic regression of small expressions}.
\newblock \bibinfo{journal}{\emph{Evolutionary Computation}} \bibinfo{volume}{29}, \bibinfo{number}{2} (\bibinfo{year}{2021}), \bibinfo{pages}{211--237}.
\newblock


\end{thebibliography}
\end{document}